# A Novel Approach for Intelligent Robot Path Planning

TIRTHARAJ DASH
Department of Information Technology
National Institute of Science and Technology
Berhampur-761008, India
Email: tirtharajnist446@gmail.com

GOUTAM MISHRA
Department of Electrical and Electronics Engineering
National Institute of Science and Technology
Berhampur-761008, India
Email: goutamishra@gmail.com

TANISTHA NAYAK
Department of Information Technology
National Institute of Science and Technology
Berhampur-761008, India
Email: tanisthanist213@gmail.com

*Abstract—* **Path planning of Robot is one of the challenging fields in the area of Robotics research. In this paper, we proposed a novel algorithm to find path between starting and ending position for an intelligent system. An intelligent system is considered to be a device/robot having an antenna connected with sensor-detector system. The proposed algorithm is based on Neural Network training concept. The considered neural network is Adaptive to the knowledge bases. However, implementation of this algorithm is slightly expensive due to hardware it requires. From detailed analysis, it can be proved that the resulted path of this algorithm is efficient.**

## I. INTRODUCTION

Over the last decade, **Intelligent** robots are driven by complex control systems. Different modules in these control system provides different control task. Major module in the intelligent robot or vehicle is the path planning module. The concept becomes more challenging in the field of path planning if the path and the environment are totally unknown. However, path planning refers to deriving a shortest path for the robot's movement. Different methods and algorithms have been proposed in this area [1]. Evolutionary and meta-heuristic algorithms have been developed and used as search and optimization tools in various problem domains, including science, commerce, and engineering. Their broad applicability, ease of use, and global perspective may be considered as the primary reason for their success. The Swarm-based approach to optimization has a great role on the intelligent robot path planning. When selecting the best subset of observation locations subject to constrained resources viz. limited time or battery capacity it is an important problem to trade off exploration like gathering information about the environment and exploitation using the current knowledge about the environment effectively for efficiently. Even the non-adaptive setting, where paths are planned before observations are made, is NP-hard, and has been subject to much research [1,4].

Recently, there has been an increase in demands to enhance intelligence of automation systems for construction operations in hazardous work environments such as underwater, in chemically or radioactively contaminated areas, and in regions with harsh temperatures. An automated construction system consists of several pieces of hardware like mobile robots and software for operating these robots. In developing an intelligent mobile construction robot, a navigation system that can produce an effective and efficient path-planning algorithm is a very important element for this purpose. The purpose of a path-planning method for a mobile construction robot is to find a continuous collision-free path from the initial position of the construction robot to its target position. Another example like, a mobile robot could be used to transport details and sub-assemblies between a store and production lines; this task implies repeated traversal between the store and the production cells. A mobile robot can also be used for surveillance; this task implies visiting certain checkpoints on a closed territory in a predefined order [2, 3].

The goal of this work is to demonstrate some works related to intelligent robot path planning. The methods are described in brief below. The source of these works is mainly Google Scholar. Finally, we have given a demonstration of our proposed algorithm for the same purpose based on Adaptive Neural Network (ANN).

## II. SURVEY

### A. Adaptive Constrained Distance Transformation

Horng and Li proposed a method to solve the problem of vehicle path planning. Their technique was based on the adaptive constrained distance transformation. This technique uses incremental distance, vehicle characteristics, spatial properties of the terrain data etc. and later on this is interpreted in terms of incremental cost. As claimed, this technique of path planning proved to be far more efficient than the Euclidean distance and constraint based techniques [1]. The output of this method is given in Figure-1 below. Here the black area resembles with the considered robot or vehicle and bar is the obstacle to the robot. So, the robot has to travel without colliding with the obstacle.

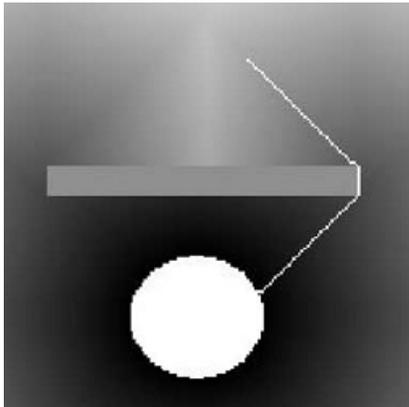

Figure-1: Output of Horng and Li's Technique [1]

### B. UAV Tracking Method

Lee et al. had also contributed to a method closely related to the solution for vehicle path planning problem. Their work was based on strategies of path planning for UAV to track a ground vehicle. In this technique it has been stated that the ground vehicle's motion dynamics is arbitrary at any time instant whereas on the contrary the motion dynamics of the UAV is fixed and at all time instants it has to maintain its motion parameters for an efficient tracking performance. Their algorithm has taken wind disturbance as one of the key parameters for causing off-course in direction of UAV's motion and has rendered an effective solution to counteract with such disturbances during any in flight operation [2]. The algorithm can be viewed using the figure below.

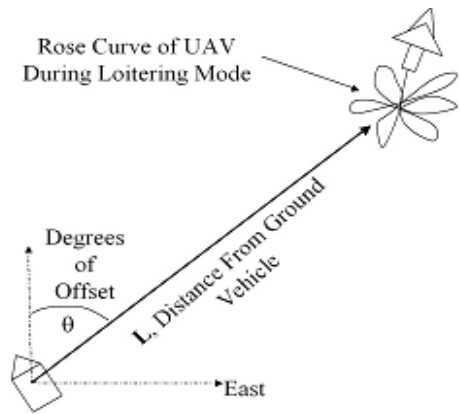

Figure-2: UAV Tracking algorithm [2]

### C. Scheduled Optimization for Multi-Robot System

LaVlle and Hutchinson, in their work towards contribution to path planning problem solutions proposed optimized schedules for a prioritized path planning in the case of Multi Robot Systems. The work claimed its efficiency over conventional prioritized and decoupled planning techniques by basing its importance by implementation of optimized priority schemes.

These schemes in this work have stated that they have aimed to reduce path length and determination of hill climbing through random search [3]. The figure-3 shows the result of this work.

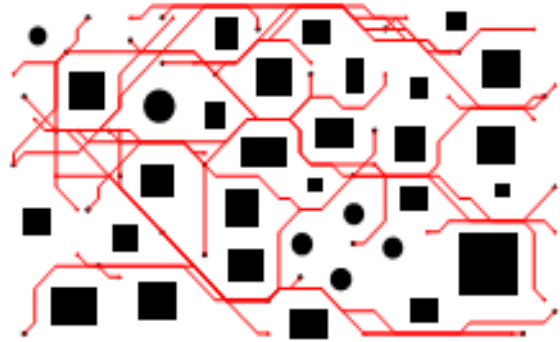

Figure-3: Result of scheduled optimization

### D. Personal Robot Planning

Rowe and Alexander in their work towards contribution to path planning have provided a description regarding path planning of a personal robot. In their work they have provided description of the personal robot and have also mentioned regarding a type of path planner which uses a global path potential field approach which has also been accompanied with an optimization criterion. This technique of their boasted that using this technique the robot maintains safe distance from the obstacles and simultaneously obtaining an optimal path by using stored partial knowledge about terrain data. [4]

### E. Contemporaty Logic Design

Katz and Borriello have provided a description regarding usage of search based algorithms and discrete environments for mobile robot's path planning. In this technique they have implemented Breadth for Search in both hardware and software environments. This technique claimed that when tested the time gain in hardware was more than software and thus stated that due to this fact along with the advantage of hardware reconfigurability multiple tasks could be implemented with a possibility to use suitable task befitting for the application. [5]

### F. Robot Path planning for Earth-work operation

Kim et al. in their work have described about a navigation system for intelligent mobile construction robot. Their work claimed that the navigation system devised by them has been implemented with an improvised Bug based algorithm named Sensbug, which is capable of producing an effective path in an unknown environment which involved both stationary and movable obstacles. [6]

### G. Trajectory Generation Method

Lian and Murray in their work have described about Cooperative Path Planning design methodology for multi vehicle systems and a Non linear trajectory generation algorithm. In this work they have taken examples of three scenarios of multivehicle tasking in CPP framework and have used the Non Linear Generation algorithm for production of real time trajectory based upon desired vehicle characteristics. [7]. Figure-4 shows the methodology of this work.

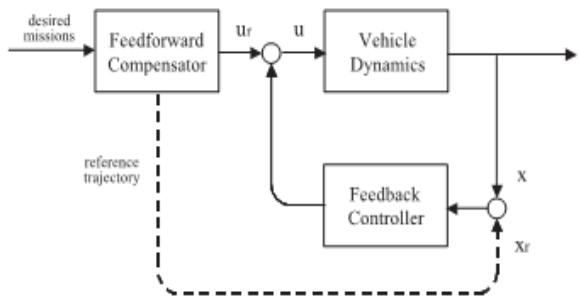

Figure-4: Trajectory Method [7]

*H. Ontology based method*

Provine et al. in their work towards contribution towards vehicle path planning have used an ontology that supports reasoning out obstacles for improvisation of capabilities and performance of onboard route planning in autonomous vehicles. In their work they have also used rules that act in conjunction with the ontology that's capable of estimating the damages that incurs from collision against different objects in different environments. The implementation of this automated reasoning helps in deciding that whether to plan or to avoid the obstacles. [8]

*I. Optimal path planning for known environment*

Stentz in his work towards to contribution to vehicle path planning has used an effective algorithm named D* which is capable of generating path planning in unknown, partially unknown and changing environments in optimal and complete manner. The work claims to resolve issues in an exploratory robot by the help of this algorithm. [9]

*J. Path planning for Deformation robots*

Gayle et al. in their work have represented a path planning algorithm for flexible robots in complex environments. The work boasted that the algorithm is able to compute a collision free path by taking into consideration of various parameters into account such as various geometric and physical constraints, obstacle avoidance, non-penetration constraint, volume preservation, surface tension energy preservation etc.. The work has also claimed that the algorithm has also been implemented in practical environments such as path planning of catheters in liver chemoembolization. [10]

*K. Honey bee mating optimization algorithm*

Sahoo et al. in their work towards contribution to vehicle path planning have implemented Honey Bee Mating Optimization technique for Navigational Planning in the case of Multi Robot environment. Their work has also claimed that this technique has exhibited the same performance as that of prior evolutionary techniques and so also the work also claims certain level of high better performance than the previous prior techniques [11].

*L. Use of Probability Recursive Function*

Khalil and Kazem in their work towards vehicle path planning have stated an efficient path planning software modeling technique using Beziers curve method for mobile robot path planning. The work has claimed that has improvised the Beziers curve method approach by overcoming its major drawbacks. The work has stated regarding the implementation of the Up and down direction and the Probability Recursive Function (PRF) techniques to overcome the drawbacks. The PRF technique has claimed that it has been successful by making a robot traverse from its starting point to end point with full percent obstacle avoidance and selection of optimal path without making any comparison between different paths [12]. However, the theory of computation can be applied to save the state of the robot at different time. For this purpose, an efficient computing model viz. Quantum Turing Machine (QTM) can be used [13, 14].

The next section shows a brief description of our proposed algorithm for intelligent robot path planning based on the concept of Adaptive Neural Network (ANN). However, Adaptive Resonance Theory can be applied for this purpose also [15].

### III. PROPOSED ALGORITHM

We have used Adaptive Neural Network to be trained by the state and position of the robot. The position is considered to be the pixel value of the current position or target position.

A diagram is given in Figure-5 below to show the demonstration.

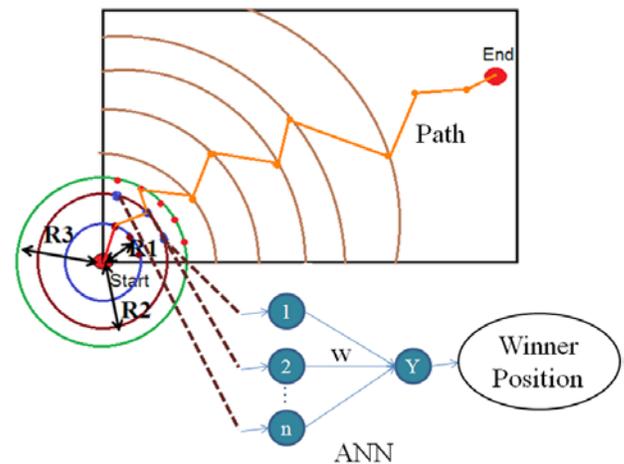

Figure-5: Proposed Methodology using ANN

However, the robot should have a sensor and detector system attached to its antenna as given in Figure-6.

**Algorithm:**
1. Put the Robot in its starting position
2. Set the Finishing Point (End)
3. Set a boundary for the robot
4. For each sliding (distance x)

5. Do
6. Rotate the rotating bar at angle $a^0$
7. Sense the point corresponding to the angle 'a'.
8. Feed the point to the ANN
9. Use Steps 6-8 for the limit of the boundary.
10. End Step-4.
11. Process the ANN operation to select the winner based on the interlayer weight (W) of the ANN.
12. Selected winner is the Next position of the Robot.
13. If (Next Position == Finish)
14. STOP
15. Else Goto-16.
16. Repeat Steps 4-12 to get the Next Position.

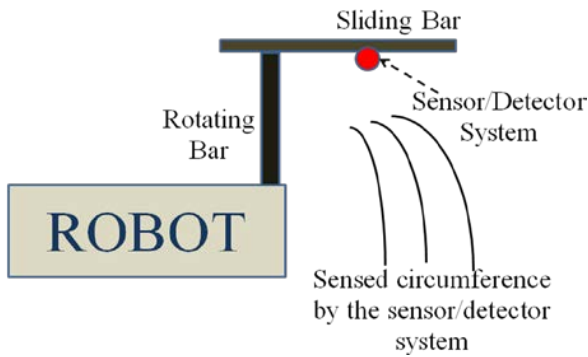

Figure-6: Hardware Organization of the Robot for this method

The basic criterion for application of this approach is when the considered area is homogeneous with respect to color. So, that the sensor can sense the obstacle but will not use that position to train the network.

### IV. CONCLUSION.

In this work, we proposed a novel algorithm for intelligent path planning. But for the implementation purpose, expensive hardware is required. The advantages of this method are that the finding the next position of the robot is fast due to use of the Neural Network concept and use of Sensor/Detector system. However, this algorithm may not guarantee a shortest path, but it will prove itself to be the best to give an efficient path. Obviously, for future work, implementing such a hardware to test the efficiency of this algorithm will be crucial to the area of Robotics and Embedded System.